\documentclass[runningheads]{llncs}
\usepackage{graphicx}

\usepackage{multirow}
\usepackage{bm}
\usepackage{amsmath,amssymb,amsfonts}
\usepackage{graphicx}
\usepackage{booktabs}
\usepackage{balance}
\usepackage[dvipsnames]{xcolor}
\usepackage{subcaption}
\usepackage{enumerate}
\usepackage{makecell}
\usepackage[misc]{ifsym}
\usepackage{hyperref}
\hypersetup{hidelinks=true}

\definecolor{darkgreen}{RGB}{78,167,46}
\definecolor{lightblue}{RGB}{15,158,213}

\begin{document}
\title{Fine-grained Prompt Tuning: A Parameter and Memory Efficient Transfer Learning Method for High-resolution Medical Image Classification}
\titlerunning{Fine-grained Prompt Tuning}
%
\author{Yijin Huang$^{1,2}$ \and
 Pujin Cheng$^{1,3,4}$ \and
 Roger Tam$^{2(\textrm{\Letter})}$ \and
 Xiaoying Tang$^{1,4(\textrm{\Letter})}$}
 %
 
 \authorrunning{Y. Huang et al.}
%
\institute{ Department of Electronic and Electrical Engineering, \\ 
Southern University of Science and Technology, Shenzhen, China \\
\email{tangxy@sustech.edu.cn} \and
School of Biomedical Engineering, \\
The University of British Columbia, Vancouver Canada \\
\email{roger.tam@ubc.ca} \and
Department of Electrical and Electronic Engineering, \\
The University of Hong Kong, Hong Kong, China \and
Jiaxing Research Institute, \\
Southern University of Science and Technology, Jiaxing, China
}
%
\maketitle              
\begin{abstract}
Parameter-efficient transfer learning (PETL) is proposed as a cost-effective way to transfer pre-trained models to downstream tasks, avoiding the high cost of updating entire large-scale pre-trained models (LPMs). In this work, we present Fine-grained Prompt Tuning (FPT), a novel PETL method for medical image classification. FPT significantly reduces memory consumption compared to other PETL methods, especially in high-resolution input contexts. To achieve this, we first freeze the weights of the LPM and construct a learnable lightweight side network. The frozen LPM takes high-resolution images as input to extract fine-grained features, while the side network is fed low-resolution images to reduce memory usage. To allow the side network to access pre-trained knowledge, we introduce fine-grained prompts that summarize information from the LPM through a fusion module. Important tokens selection and preloading techniques are employed to further reduce training cost and memory requirements. We evaluate FPT on four medical datasets with varying sizes, modalities, and complexities. Experimental results demonstrate that FPT achieves comparable performance to fine-tuning the entire LPM while using only 1.8\% of the learnable parameters and 13\% of the memory costs of an encoder ViT-B model with a 512 $\times$ 512 input resolution.
\end{abstract}
\begin{figure}[t]
\begin{multicols}{2}[\columnsep=0.5cm]
    \centering
    \includegraphics[width=0.9\columnwidth]{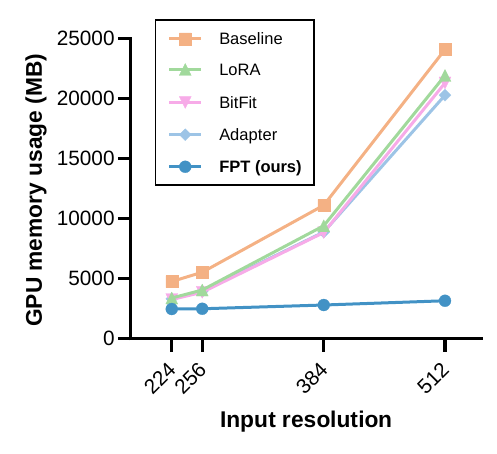}
    \caption{High-resolution comes at the cost of heightened GPU memory consumption.}
    \label{memory}
\columnbreak
    \centering
    \includegraphics[width=\columnwidth]{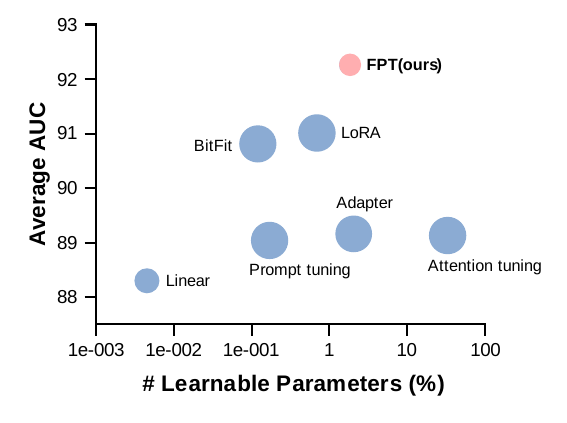}
    \caption{Our proposed FPT shows the best trade-off between performance and efficiency. The size of the dots represents memory usage.}
    \label{tradeoff2}   
\end{multicols}
\vspace{-0.5cm}
\end{figure}

\section{Introduction}
By utilizing the technique of fine-tuning \cite{weiss2016survey}, pre-trained models can be effectively adapted to specific downstream tasks by initializing the task-specific model with the weights from the pre-trained model. Recently, the remarkable achievements of large-scale pre-trained models (LPMs) \cite{devlin2018bert,kirillov2023segment,radford2021learning,radford2019language} further underscore the importance of this technique. However, as the model size grows rapidly, fine-tuning the parameters of an entire LPM has become very costly. To address this challenge, the concept of parameter-efficient transfer learning (PETL) \cite{houlsby2019parameter,hu2021lora,jia2022visual,sung2022lst,wu2023medical,zhang2020side} has emerged, offering a strategic approach for transferring pre-trained models. PETL involves selectively updating a small subset of pre-trained parameters or introducing a modest number of additional parameters specific to new tasks, while keeping the majority of pre-trained parameters frozen.

PETL has successfully established its effectiveness in computer vision, but the field of medical image has not fully benefited from such advances yet \cite{dutt2023parameter}, mainly because of the domain gap between them. In natural images, the objects of interest typically occupy a large portion of the image and exhibit distinct characteristics. In contrast, diagnostic cues in medical images often occupy a small portion and are distributed throughout the entire image \cite{huang2024ssit}. Providing such fine-grained information often requires the use of high-resolution input images \cite{chen2021super,huang2023identifying,zhang2023prompt}. However, as shown in Fig. \ref{memory}, this preference for high-resolution images comes at the cost of increased GPU memory consumption and training expenses.

In this work, we propose a novel parameter and memory efficient transfer learning method, namely Fine-grained Prompt Tuning (FPT). FPT aims to enhance the effectiveness of PETL specifically for medical images in high-resolution input contexts by addressing two main challenges:

\textbf{\textit{1) How to efficiently extract fine-grained information from high-resolution images?}} Existing PETL methods involve training a subset of parameters within the large-scale pre-trained model (LPM) \cite{houlsby2019parameter,hu2021lora}. In contrast, FPT utilizes a lightweight additive network inspired by the concept of a side network \cite{sung2022lst,zhang2020side}. This learnable side network is introduced outside the LPM, eliminating the need for back-propagation through the LPM. However, training the side network can still be computationally expensive with high-resolution input images due to the long input sequence. FPT addresses this concern by strategically introducing asymmetric input resolution and employing important token selection to significantly reduce the length of the input sequence for the learnable network.

\textbf{\textit{2) How to effectively adapt pre-trained knowledge from LPMs?}} LPMs are primarily pre-trained on natural image datasets like ImageNet \cite{ridnik2021imagenet}. To effectively adapt pre-trained knowledge from LPMs of domains outside of medical images, FPT introduces the concept of fine-grained prompts and a Fine-grained Fusion Module (FFM) as bridging components. Fine-grained prompts are a small set of learnable embeddings that summarize pre-trained knowledge from the LPM through the FFM. These prompts are then prepended to the intermediate layers of the side network to convey fine-grained information by integrating them into the forward propagation.

Our main contributions are summarized as follows:
\begin{enumerate}
    \item We present a novel PETL method, namely Fine-grained Prompt Tuning (FPT), for medical image classification in high-resolution contexts. Asymmetric input and important token selection are proposed to improve memory efficiency. Fine-grained prompts and fine-grained fusion module are introduced to adapt pre-trained knowledge effectively and efficiently. Our code is available \href{https://github.com/YijinHuang/FPT}{\textcolor{blue}{online}}.
    \item To the best of our knowledge, this is the first work to enhance the efficiency of PETL in high-resolution input settings, which is particularly significant in the field of medical image analysis.
    \item We introduce a new metric to evaluate the trade-off between performance and memory efficiency for transfer learning methods.
    \item We conduct extensive experiments on four medical image datasets with different modalities. As shown in Fig. \ref{tradeoff2}, FPT achieves the best trade-off between performance and parameter/memory efficiency.
\end{enumerate}

\section{Method}
\subsection{Side Tuning}
\begin{figure}[t]
	\centering
	\includegraphics[width=0.92\textwidth]{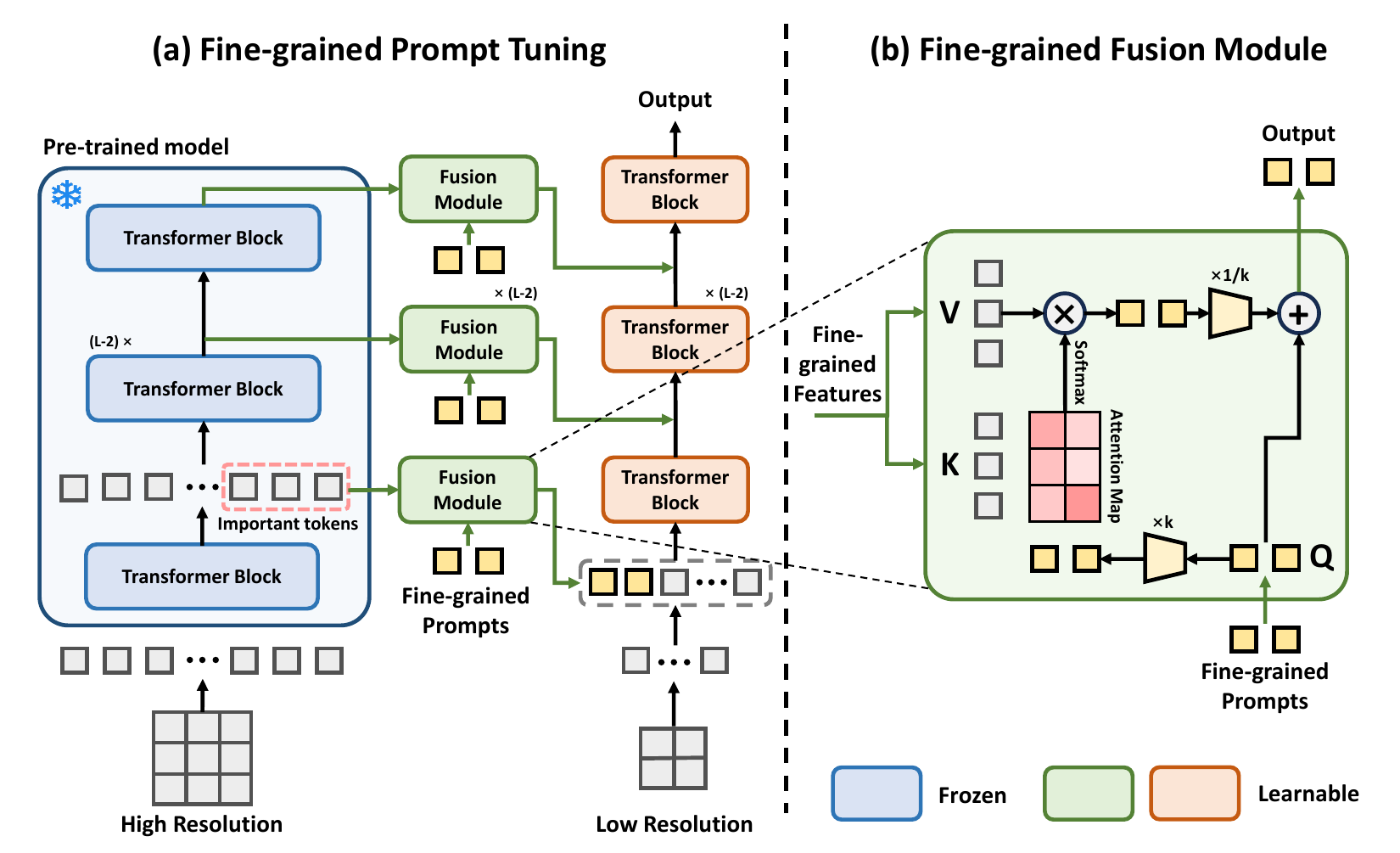}
	\caption{The proposed FPT framework.}
	\label{framework}
\end{figure}

As illustrated in part (a) of Fig. \ref{framework}, the FPT framework consists of two networks: a frozen LPM $M$ and a learnable side network $S$. Unlike other PETL methods that introduce additional learnable parameters within the LPM, in our approach, the entire LPM remains frozen while the side network is kept learnable and separate. We adopt a lightweight architecture design for the side network, which is a scaled-down variant of the LPM. The hidden dimensions of the side network are $1/k$ times that of the LPM, where $k$ represents a reduction factor. To leverage the pre-trained knowledge from the LPM, the side network reuses the intermediate features at each layer of the LPM. Specifically, given two models $M$ and $S$ with $L$ layers, parameterized by $\theta_M$ and $\theta_S$ respectively, the intermediate activation $z_{S,\mathrm{I}}^l$ of layer $l$ is obtained as follows:
\begin{align}
&z_M^l = \theta_M^l (z_M^{l-1}), \\
&z_S^l = \theta_S^l (\mathcal{F}(z_S^{l-1}, z_M^{l})),
\end{align}
where $\mathcal{F}$ denotes the module that fuses features of $M$ and $S$, and $z_{S,I}^L$ is considered the final output of the framework. The use of the side network not only reduces the number of trainable parameters due to its lightweight design but also helps mitigate memory expenses during the training phase. As shown in part (a) of Fig. \ref{modules_combined}, the side network eliminates the need for resource-intensive back-propagation from the pre-trained model by excluding any learnable parameters in the forward pass of the heavy LPM model.

\subsection{Asymmetric Input}
Fine-grained information holds significant importance in the context of medical image analysis and is typically acquired by using high input resolutions. Although training with high-resolution inputs may be infeasible due to high memory consumption, using a LPM solely for inference with high-resolution images remains practical. Therefore, we propose an asymmetric input strategy within the FPT framework. Specifically, given a high-resolution image $I$, we simply resize it to obtain a low-resolution image $I^\prime$. Then, the frozen pre-trained model $M$ is provided with the image $I$, while the learnable side network $S$ is fed low-resolution image $I^\prime$. Thus, we have the intermediate activation of the side network $z_{S,I^\prime}^l = \theta_S^l (\mathcal{F}(z_{S,I^\prime}^{l-1}, z_{M,I}^{l}))$.

\subsection{Fine-grained Prompts and Fusion Module}
In this section, we introduce our proposed fusion module $\mathcal{F}$. As shown in the part (b) of Fig. \ref{framework}, we utilize the cross-attention mechanism \cite{chen2021crossvit} inside the FFM to fuse features from the LPM to the side network. In the context of cross-attention, we reuse the key and value from the self-attention layer of the pre-trained model $M$. Regarding the query, one approach is to directly reuse the query from the side network $S$. However, the cross-attention map can be large if the input sequence is long, leading to increased memory consumption. Therefore, we introduce a small set of learnable embeddings $z_p$, namely fine-grained prompts, into each layer of the fusion modules as the query. Unlike prompt tuning \cite{jia2022visual,lester2021power}, which directly uses prompts as part of the input sequence, fine-grained prompts serve as a bridge linking the frozen LPM and the side network. These prompts are concatenated with the intermediate sequence of the side network to join the forward propagation after fusing pre-trained features from the LPM. They are then removed after the layer's forward processing. Specifically, the fusion module is processed as follows:
\begin{align}
&\mathcal{F}(z_S, z_M) = [z_S, f_{out}(\text{CA}(z_p, z_M)) + z_p], \\
&\text{CA}(z_p, z_M) = \text{Attn}(Q, K, V) = \text{Attn}(f_{in}(z_p), K_M(z_M), V_M(z_M)),
\end{align}
where the notations for the input source and the index of the layer are omitted in the formula for simplification. Here, $\text{CA}(\cdot)$ and $\text{Attn}(\cdot)$ denote the cross-attention module and attention function \cite{vaswani2017attention} respectively, $[\cdot, \cdot]$ denotes the concatenation operation, $f_{in}$ and $f_{out}$ denote linear layers that align the hidden dimension between the features. The terms $K_\mathrm{M}$ and $V_\mathrm{M}$ refer to the key and value mapping matrices within the corresponding self-attention layer of $M$ respectively.

\begin{figure}[t]
	\centering
	\includegraphics[width=\textwidth]{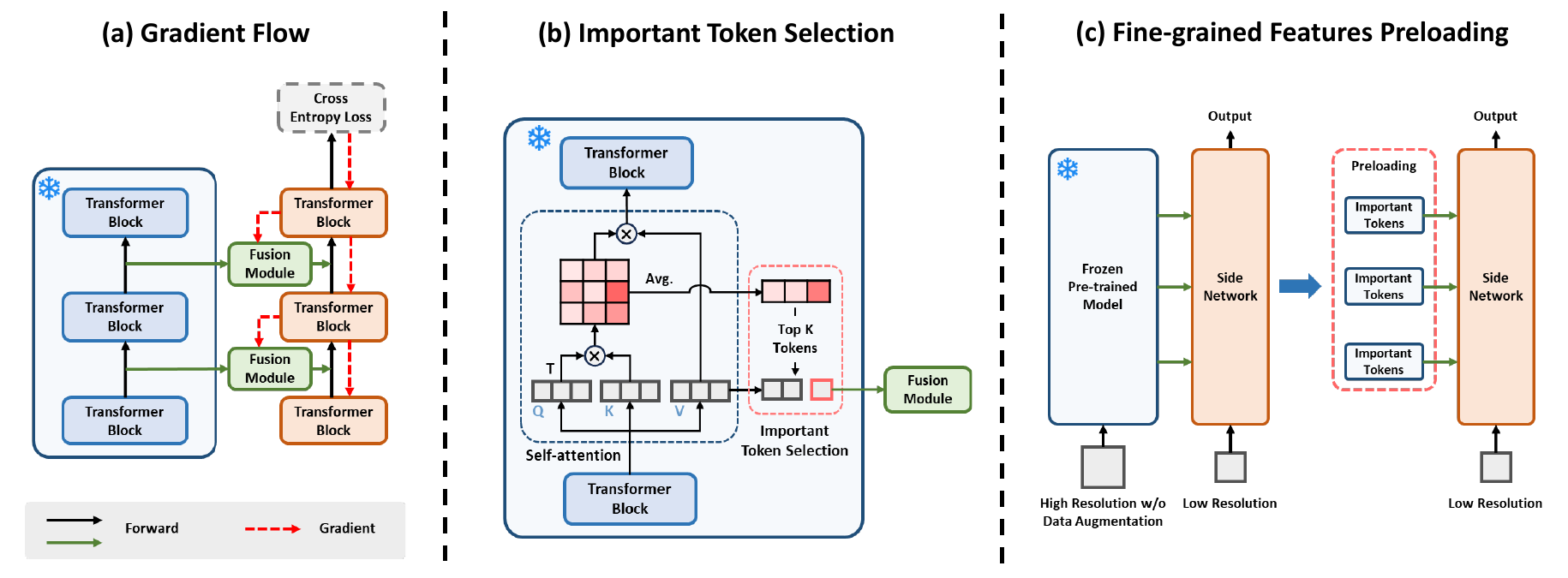}
	\caption{
		Components of FPT: (a) The gradient flow of FPT. (b) Important tokens selection mechanism. (c) Fine-grained features preloading.}
	\label{modules_combined}
\end{figure}

\subsection{Important Token Selection}
Building on knowledge in the medical image domain, where images of the same modality often display similar anatomical structures and the objects of interest typically occupy a small proportion of the entire image, we propose a method to further reduce memory consumption, as shown in part (b) of Fig. \ref{modules_combined}. Specifically, we introduce important token selection, which selects the top $m\%$ tokens with the highest average scores on the self-attention map, considering them as important tokens. Only the features associated with these important tokens are passed to the FFM. This approach significantly reduces the overhead introduced by high-resolution inputs while preserving essential fine-grained information.

\subsection{Fine-grained Features Preloading}
To further accelerate training procedure, we opt not to use any data augmentation on the input of the frozen LPM. This choice ensures that the intermediate features from the LPM associated with an image remain consistent throughout training. This approach allows us to pre-store these features before training, leading to significant reductions in training costs. Note that while the preloaded features remain fixed, the data augmentation applied to the input of the side network maintains the diversity of training samples.

\section{Experiments}
\subsection{Datasets}
We evaluate FPT on four medical datasets with different modalities, including fundus images (messidor-2 \cite{messidor}), dermoscopic images (ISIC 2018 \cite{codella2019skin}), mammography (DDSM \cite{lee2017curated}), and chest X-ray (COVID \cite{Siddhartha_2021}). The dataset sizes range from 1,748 to 11,527 samples, with classification categories varying from 3 to 7. We use official dataset splits when available. Otherwise, we employ a random partition of 70\%/10\%/20\% for training, validation, and testing, respectively.

\subsection{Training and Evaluation Setup}
\subsubsection{Experiment setup}

In this study, we utilize a popular variant of Vision Transformers (ViT) \cite{dosovitskiy2020image}, specifically ViT-B(ase), which is pre-trained on ImageNet-21K \cite{ridnik2021imagenet}. All methods are fine-tuned for 20 epochs with a mini-batch size of 16. We use the AdamW optimizer \cite{loshchilov2017decoupled} with cross-entropy as the loss function for all datasets. To ensure fair comparisons, we conduct a grid search of hyper-parameters for all methods. All methods are run at a resolution of $512 \times 512$. In the context of FPT, the high input resolution for the LPM remains $512 \times 512$, while the low input resolution for the side network is set to $224 \times 224$. All methods were trained employing the same data augmentations as those for low-resolution inputs in FPT. We set the reduction factor $k$ to 8 and use 16 fine-grained prompts with the same hidden dimension as $S$. For important token selection, we retain the top 20\% of important tokens.

\subsubsection{Evaluation metric}
We use the Area Under the Receiver Operating Characteristic Curve (AUC) to evaluate the classification performance of each dataset. The performance-efficiency (PE) metric \cite{he2022parameter,li2022elevater} is a metric to assess the performance and efficiency trade-off. However, PE only considers the impact of the number of learnable parameters. The memory requirement is another crucial factor that significantly influences training expenses. Therefore, we extend the PE metric to include performance-parameter-efficiency (PPE) and performance-memory-efficiency (PME). The PPE formula remains the same as PE, defined as $\text{PPE} = \text{score} * \exp(-\log_{10}(r + 1))$, where the score represents the average performance across all datasets, and $r$ is the ratio of learnable parameters to all parameters. Similar to PPE, PME is defined as $\text{PME} = \text{score} * \exp(-\log_{10}(m + 1))$, where $m$ is the ratio of the GPU memory requirement of the method to that of fine-tuning the entire LPM.

\begin{table}[t]
    \centering
    \renewcommand{\arraystretch}{1.2}
    \caption{Comparison results with state-of-the-art PETL methods across four evaluation datasets. `Params.' denotes the ratio of learnable parameters to the total number of parameters. `Mem.' denotes the memory usage (MB). The best and second-best results are bold and underlined, respectively.}
    \resizebox{\textwidth}{!}{%
    \begin{tabular}{lccccccccc}
    \toprule
    \multirow{2}{*}{Method}            & \multicolumn{2}{c}{Computing cost}                & \multirow{2}{*}{\begin{tabular}[c]{@{}c@{}}Fundus\\ (Messidor2)\end{tabular}} & \multirow{2}{*}{\begin{tabular}[c]{@{}c@{}}Dermoscopy\\ (ISIC2018)\end{tabular}} & \multirow{2}{*}{\begin{tabular}[c]{@{}c@{}}Mammography\\ (DDSM)\end{tabular}} & \multirow{2}{*}{\begin{tabular}[c]{@{}c@{}}Chest X-ray\\ (COVID)\end{tabular}} & \multicolumn{3}{c}{Performance}                              \\ \cline{2-3} \cline{8-10} 
                                       & Params.                  & Mem.                   &                                                                                     &                                                                                         &                                                                               &                                                                                & Avg. AUC                & PPE                      & PME                      \\ \hline
    \textcolor{gray}{Full fine-tuning} & \textcolor{gray}{100}    & \textcolor{gray}{24,116}& \textcolor{gray}{86.87 $\pm$ 0.53}& \textcolor{gray}{96.65 $\pm$ 0.29}& \textcolor{gray}{92.49 $\pm$ 0.34}& \textcolor{gray}{99.85 $\pm$ 0.05}& \textcolor{gray}{93.96}& \textcolor{gray}{69.54}& \textcolor{gray}{69.54}\\
    \textcolor{gray}{Linear probing}   & \textcolor{gray}{0.01} & \textcolor{gray}{4,364}& \textcolor{gray}{79.73 $\pm$ 0.84}& \textcolor{gray}{93.37 $\pm$ 0.31}& \textcolor{gray}{80.89 $\pm$ 0.39}& \textcolor{gray}{99.21 $\pm$ 0.07}& \textcolor{gray}{88.30}& \textcolor{gray}{88.30}& \textcolor{gray}{82.15}\\
    Prompt tuning \cite{jia2022visual}                     & 0.17                     & 21,530& 80.02 $\pm$ 2.34& 94.20 $\pm$ 0.20& 82.67 $\pm$ 0.34& 99.27 $\pm$ 0.03& 89.04& 88.97& 67.49\\
    Attention tuning \cite{touvron2022three}                  & 33.04                    & 21,740& 81.87 $\pm$ 1.42& 94.40 $\pm$ 0.40& 80.67 $\pm$ 2.33& 99.58 $\pm$ 0.19& 89.13& 78.74& 67.42\\
    Adapter \cite{houlsby2019parameter}                           & 2.05                     & 20,308& 80.77 $\pm$ 1.48& \textbf{95.96 $\pm$ 0.13}& 80.76 $\pm$ 3.38& 99.17 $\pm$ 0.48& 89.16& 88.38& 68.39\\
    BitFit \cite{zaken2021bitfit}                            & 0.12                     & 21,330& 83.81 $\pm$ 1.11& 94.84 $\pm$ 0.15& \underline{84.77 $\pm$ 0.54}& \textbf{99.81 $\pm$ 0.03}& 90.81& \underline{90.76}& \underline{68.97}\\
    LoRA \cite{hu2021lora}                              & 0.69                     & 21,944& \textbf{86.08 $\pm$ 0.95}& \underline{95.02 $\pm$ 0.22}& 82.26$\pm$ 4.04& 99.69 $\pm$ 0.05& \underline{91.01}& 90.74& 68.72\\
    FPT (Ours)                         & 1.81                     & 3,182                   & \underline{84.95 $\pm$ 2.01}& 93.88 $\pm$ 0.60                                                                        & \textbf{90.52 $\pm$ 0.59}& \underline{99.70 $\pm$ 0.30}& \textbf{92.26}& \textbf{91.54}& \textbf{87.42}\\ \bottomrule
    \end{tabular}%
    }
    \label{main}
\end{table}

\subsection{Comparisons with State-of-the-art}
We compare FPT against full fine-tuning, linear probing, and state-of-the-art (SOTA) PETL approaches. In full fine-tuning, all parameters of the LPM are made learnable during training on the downstream tasks. Linear probing involves solely training the new task-specific head on top of the LPM. Generally, full fine-tuning often represents the upper performance bound for transfer learning, while linear probing represents the lower bound. We also compare FPT against five other popular SOTA PETL methods.

The performance and efficiency of PETL methods are tabulated in Table \ref{main}. It can be observed that fine-tuning the entire LPM (full fine-tuning), as the upper bound, achieves the best average AUC of 93.96\%, requiring 24,116MB of training GPU memory with a batch size of 16. Although other PETL methods improve transfer learning efficiency by reducing the number of learnable parameters, they are unable to reduce the overhead brought by the high input resolution; all compared PETL methods require at least 20,000MB of memory, and even linear probing requires 4,364MB of memory. In contrast, FPT achieves the second-best AUC of 92.26\%, while significantly reducing the memory requirement to only 3,182MB (13\% of full fine-tuning). Moreover, with a separate network, FPT utilizes only 1.81\% learnable parameters compared to full fine-tuning. Therefore, in terms of efficiency, FPT achieves the best PPE and PME, presenting the best trade-off between performance and efficiency. This parameter and memory efficiency positions FPT as a practical and feasible choice for leveraging LPMs in high-resolution contexts.

\subsection{Impact of Components}
To assess the impact of the proposed components in FPT, we evaluate the performance and efficiency of our framework by incrementally incorporating components. As shown in Table \ref{ablation}, starting with a sole side network, although the side network is lightweight, the long input sequence still consumes a large amount of memory at 17,218MB for training. Employing fusion modules with fine-grained prompts to extract pre-trained knowledge from a frozen LPM notably enhances performance but further increases the memory burden. Then, the introduction of asymmetric input significantly lowers memory usage by 58\% through decreasing the resolution of the input for the side network. It is worth noting that decreasing the input resolution for the side network enhances performance because the fine-grained information provided by the fine-grained prompts is sufficient, and smaller inputs for the side network reduce redundancy in features. Finally, by applying important token selection and preloading techniques, FPT further lowers the memory requirement by 64\% without any loss of performance.

\subsection{Impact of Important Token Selection Ratio}
We evaluate the impact of different ratios of important tokens on FPT. As shown in Table \ref{ratio}, we observe that the classification performance remains similar when the ratio is between 20\% and 50\%. We then notice that 20\% of the content of an image is sufficient for diagnosis in the tasks and modalities considered. Regarding efficiency, memory requirements increase as the ratio increases, leading to lower PME. Therefore, the ratio for important token selection is set to 20\%.
\begin{table}[t]
\begin{multicols}{2}[\columnsep=0.5cm]
    \centering
    \renewcommand{\arraystretch}{1.2}
    \caption{The performance and efficiency of FPT with different components.}
    \resizebox{\columnwidth}{!}{%
    \begin{tabular}{lllllccc}
    \toprule
    \multicolumn{5}{l}{Components}                       & Avg. AUC   & Mem.   & PME   \\ \hline
    \multicolumn{5}{l}{Sole side network}                & 80.12 & 17,218 & 62.95 \\ 
      & \multicolumn{4}{l}{+ LPM w/ FFM}                 & \makecell{90.82 \\ \textcolor{red}{($+$10.70)}} & \makecell{21,070 \\ \textcolor{red}{($+$3,852)}} & \makecell{69.14 \\ \textcolor{red}{($+$6.19)}} \\ 
      &   & \multicolumn{3}{l}{+ Asymmetric input}       & \makecell{92.14 \\ \textcolor{red}{($+$1.32)}} & \makecell{8,796 \\ \textcolor{blue}{($-$12,274)}}  & \makecell{80.50 \\ \textcolor{red}{($+$11.36)}} \\ 
      &   &      & \multicolumn{2}{l}{+ Token selection} & \makecell{92.26 \\ \textcolor{red}{($+$0.12)}} & \makecell{4,880 \\ \textcolor{blue}{($-$3,916)}}  & \makecell{84.82 \\ \textcolor{red}{($+$4.32)}} \\ 
      &   &      &         & + Features preloading       & \makecell{92.26 \\ \textcolor{gray}{($+$0.00)}} & \makecell{3,182 \\ \textcolor{blue}{($-$1,698)}}  & \makecell{87.42 \\ \textcolor{red}{($+$2.60)}} \\ \bottomrule
    \end{tabular}%
    }
    \label{ablation}
\columnbreak
    \centering
    \renewcommand{\arraystretch}{1.2}
    \caption{The performance and efficiency of FPT with different ratios for important token selection.}
    \resizebox{0.66\columnwidth}{!}{%
    \begin{tabular}{cccc}
    \toprule
    Ratio & Avg. AUC       & Mem. & PME             \\ \hline
    10\%   & 91.11          & \textbf{2,760}         & \underline{86.92} \\
    20\%   & \textbf{92.26} & \underline{3,182}         & \textbf{87.42}          \\
    30\%   & 92.20          & 3,606         & 86.79          \\
    40\%   & 91.99          & 4,020         & 86.03          \\
    50\%   & \underline{92.21}          & 4,424         & 85.71          \\
    100\%   & 92.14          & 8,796         & 80.50          \\ \bottomrule
    \end{tabular}%
    }
    \label{ratio}
\end{multicols}
\vspace{-0.4cm}
\end{table}

\section{Conclusion}
In this paper, we introduce a novel PETL method, namely Fine-grained Prompt Tuning (FPT), for medical image classification. FPT significantly reduces memory requirements, particularly in the high-resolution context that commonly used in medical image analysis. To address the challenge of high memory requirement, we first adopt the design of side tuning and enhance it with an asymmetric input strategy. We then introduce fine-grained prompts and the fine-grained fusion module to allow effective adaptation of pre-trained knowledge from an out-of-domain LPM that takes images of a different scale as input. To further reduce memory requirement, important token selection and the preloading of pre-trained features are applied. By integrating these components, our PETL method achieves superior performance across four medical datasets while maintaining the best parameter and memory efficiency.

\begin{credits}
	\subsubsection{\ackname} This study was supported by the National Key Research and Development Program of China \allowbreak{(2023YFC2415400)}; the National Natural Science Foundation of China (62071210); the Shenzhen Science and Technology Program (RCYX202\allowbreak 10609103056042); the Shenzhen Science and Technology Innovation Committee (KXFZ\allowbreak 20C20122117340001); the Guangdong Basic and Applied Basic Research (2021\allowbreak A151522\allowbreak 0131).
	
	\subsubsection{\discintname}
	The authors have no competing interests to declare that are relevant to the content of this article.
\end{credits}

\bibliographystyle{splncs04}
\bibliography{./ref.bib}

\end{document}